\documentclass[letterpaper]{article} 
\usepackage[preprint]{aaai2027}  
\usepackage[hyphens]{url}  
\usepackage{graphicx} 
\usepackage{natbib}  
\usepackage{caption} 
\usepackage{algorithm}
\usepackage{algorithmic}

\usepackage{newfloat}
\usepackage{listings}
\DeclareCaptionStyle{ruled}{labelfont=normalfont,labelsep=colon,strut=off} 
\floatstyle{ruled}
\newfloat{listing}{tb}{lst}{}
\floatname{listing}{Listing}

\usepackage{booktabs}
\usepackage{amssymb}
\usepackage{multirow}
\usepackage{array}
\usepackage{tabularx}
\usepackage{amsmath}
\title{RealCAD: Towards Real-World Image-to-CAD Reconstruction \\ under Domain Shift and Parameter Bias}
\author{
    Yihe Sun\textsuperscript{\rm 1},
    Ziyu Lu\textsuperscript{\rm 1},
    Kaihua Tang\textsuperscript{\rm 1}\corresponding,
    Xian-Sheng Hua\textsuperscript{\rm 1}\corresponding
}

\affiliations{
    \textsuperscript{\rm 1}Tongji University\\
}

\begin{document}

\maketitle

\begin{abstract}
Reconstructing editable Computer-Aided Design (CAD) models from images is essential for downstream modification, manufacturing, and design reuse. 
However, existing image-to-CAD methods are developed predominantly on synthetic renderings and face two coupled obstacles: a substantial appearance domain gap between synthetic and real images, and a previously overlooked parameter bias in widely used CAD data.
We show that the local normalization adopted by DeepCAD concentrates several geometric parameters around a few discrete values while encoding substantial information in a single scale factor. Consequently, a model can achieve deceptively high parameter accuracy by exploiting these frequent values rather than inferring geometry from the input image. 
In this paper, we propose RealCAD, a unified framework that addresses these limitations at the representation, image, and feature levels. At the representation level, we redistribute scale information to the corresponding geometric parameters, producing less concentrated parameter distributions in a shared scale space. At the image level, geometry-constrained translation converts synthetic renderings toward the real-image domain while conditioning on object contours. At the feature level, a multi-positive contrastive objective aligns representations of the same CAD model across viewpoints and image domains, enabling CAD sequence prediction from each individual view. We further introduce OpenRealCAD, comprising four-view photographs of 392 3D-printed objects paired with ground-truth command sequences. Experiments show that the revised representation substantially reduces the accuracy attainable from parameter-frequency priors, making parameter accuracy a more reliable measure of image-conditioned geometric inference. RealCAD further improves real-domain command and parameter accuracy, while retaining competitive synthetic-domain performance. The code\footnote{Code: \url{www.github.com/sunyh39/RealCAD}.} and dataset\footnote{Dataset: \url{www.modelscope.cn/datasets/yeguomao/RealCAD}.} are publicly available.

\end{abstract}


\begin{figure}[t]
    \centering
\includegraphics[width=0.45\textwidth]{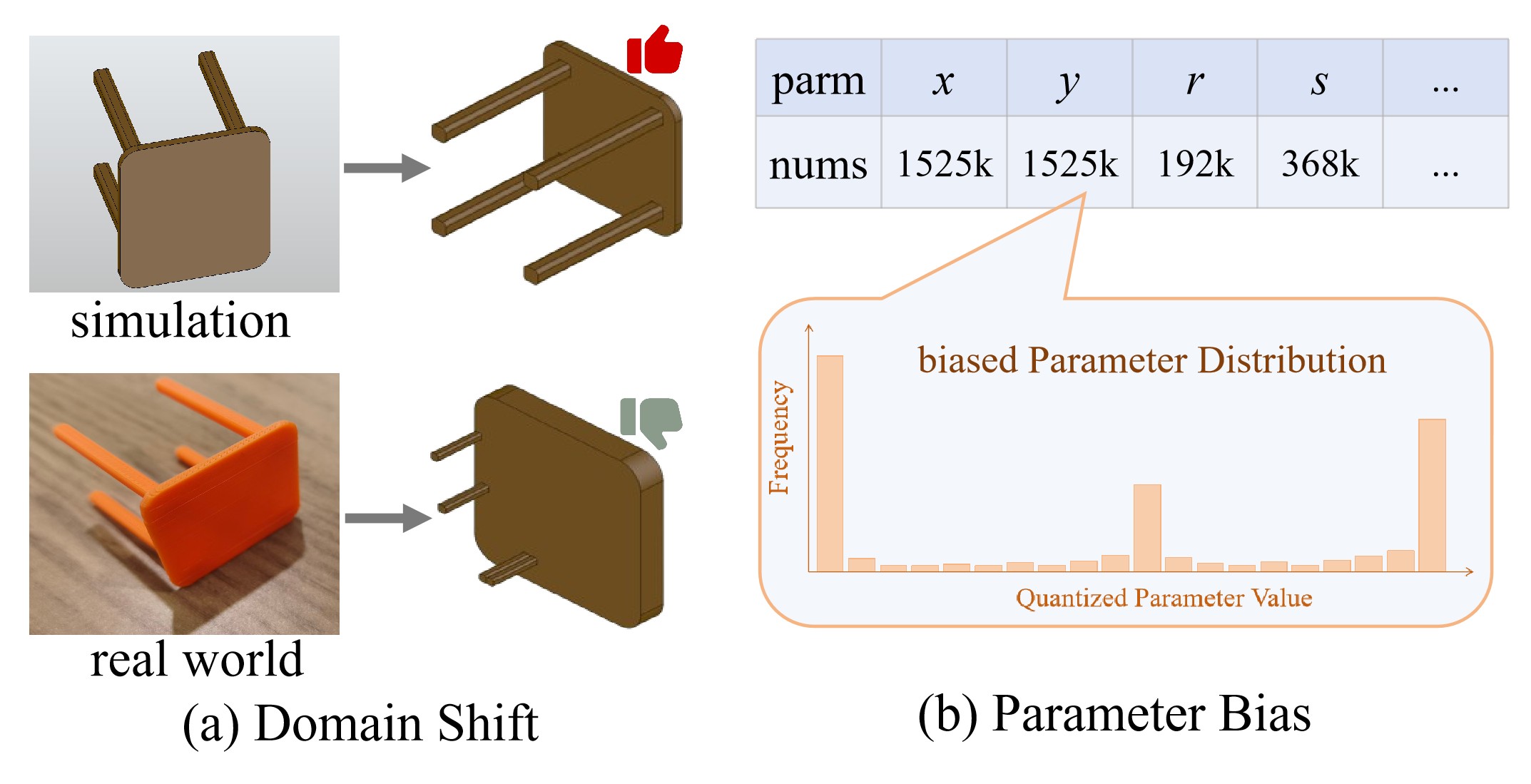}
\caption{Key challenges in image-to-CAD reconstruction.
(a) \textbf{Domain shift}: a model trained on synthetic renderings
reconstructs synthetic inputs accurately but degrades substantially
on real-world images (\textit{e.g.}, 3D printed CAD models).
(b) \textbf{Parameter bias}: the DeepCAD representation produces
highly concentrated geometric-parameter distributions,
encouraging reliance on dataset-specific shortcuts.}
    \label{DomainShift}
        \vspace{-4mm}
\end{figure}

\section{Introduction}
Image-to-CAD reconstruction aims to recover editable computer-aided design (CAD) models from images, enabling downstream modification, manufacturing, and design reuse.
Unlike surface-based representations such as meshes~\cite{groueix2018papier, nash2020polygen, wang2018pixel2mesh}
or point-based representations~\cite{achlioptas2018learning, cai2020learning, mo2019structurenet, yang2018foldingnet}, CAD models explicitly encode parametric construction histories as sequences of modeling operations.
Recent studies have explored reconstructing parametric CAD models from text, point clouds, and images. However, textual descriptions are often insufficient to specify fine-grained geometry and spatial relationships, while point-cloud-based methods typically require dense and relatively complete 3D observations. Images, by contrast, are easier to acquire and therefore provide a more practical input modality for CAD reconstruction.

However, a key challenge to practical deployment is the simulation-to-reality (Sim2Real) gap~\cite{ju2022transferring}: models trained on synthetically rendered images often exhibit substantial performance degradation when applied to real-world photographs, as illustrated in Figure~\ref{DomainShift}(a). Nevertheless, systematic Sim2Real adaptation remains underexplored in image-to-CAD reconstruction.
CADCrafter~\cite{chen2025cadcrafter} provides preliminary evidence that models trained on synthetic images can generalize to captured photographs. Yet, its real-image evaluation set is not publicly available, and the model is trained primarily on synthetic renderings without an explicit mechanism for Sim2Real adaptation. 
Beyond the visual domain gap, the underlying CAD representation itself may also hinder reliable geometric inference. We identify a previously overlooked limitation in the normalization scheme of DeepCAD~\cite{wu2021deepcad}, a widely used benchmark and representation in sequence-based CAD modeling. Specifically, DeepCAD's local normalization assigns a substantial portion of geometric scale information to a separate scale factor $S$, resulting in highly concentrated distributions for several normalized geometric parameters, as illustrated in Figure~\ref{DomainShift}(b). This imbalance may encourage models to exploit dataset-specific statistical shortcuts rather than infer the underlying geometry from visual evidence. These observations suggest that reliable robust image-to-CAD reconstruction requires addressing both visual domain discrepancy and representation-induced parameter bias.

Collecting large-scale real-image datasets with paired CAD command-sequence annotations is expensive and difficult to scale, whereas synthetic images can be automatically rendered from existing CAD models with accurate annotations and controllable acquisition conditions. A practical strategy is therefore to leverage abundant synthetic data together with limited real data. To address the representation bias and domain gaps, we revise the DeepCAD parameter representation, construct the real-image benchmark \textbf{OpenRealCAD}, and propose the Sim2Real reconstruction framework \textbf{RealCAD}.
Specifically, we redistribute the scale information encoded by DeepCAD's local normalization factor $S$ to the corresponding geometric parameters, thereby alleviating parameter concentration and reducing reliance on dataset-specific statistical shortcuts. We further construct OpenRealCAD, comprising real images of 392 3D-printed physical CAD objects captured from four fixed viewpoints under controlled conditions.

Building on the revised DeepCAD representation and OpenRealCAD, RealCAD leverages large-scale synthetic data together with limited real images to mitigate the synthetic-to-real gap at both the image and feature levels. At the image level, geometry-constrained synthetic-to-real translation adapts synthetic renderings toward the real domain while preserving geometric structure. At the feature level, multi-positive contrastive learning aligns representations of the same CAD model across viewpoints and image domains. The resulting representations are decoded into CAD command sequences and their associated parameters. Experiments show that RealCAD substantially improves reconstruction performance on real images while maintaining competitive performance in the synthetic domain.

The main contributions are summarized as follows:
\begin{itemize}
\item We identify an overlooked distributional limitation in the DeepCAD command-sequence representation and revise its parameterization by redistributing the scale information encoded by $S$ to the corresponding geometric parameters, thereby mitigating parameter concentration and frequency-based shortcut learning.
\item We construct \textbf{OpenRealCAD}, a real-image benchmark comprising 392 CAD models, four-view images of their 3D-printed physical objects, and paired ground-truth command sequences, enabling quantitative and reproducible evaluation of Sim2Real reconstruction.
\item We propose \textbf{RealCAD}, a Sim2Real image-to-CAD reconstruction framework that combines geometry-constrained synthetic-to-real translation with multi-positive contrastive learning to reduce synthetic-to-real discrepancies at both the image and feature levels.
\end{itemize}

\section{Related Work}

\subsection{Image-Based 3D Reconstruction} 
Image-based 3D reconstruction recovers 3D geometry or scene representations from one or multiple images using representations such as point clouds, meshes, voxels, neural fields, or 3D Gaussians~\cite{scharstein2002taxonomy,yao2018mvsnet,mildenhall2020nerf,kerbl20233dgaussians,Zhang_2026_CVPR}. Although these methods have achieved substantial progress in geometric reconstruction and novel-view synthesis, they primarily represent the geometry or appearance of the reconstructed object or scene and do not explicitly preserve parametric construction histories. In contrast, image-to-CAD reconstruction aims to recover executable command sequences containing modeling operations and parameters, thereby supporting history-aware editing and design reuse.

\subsection{Parametric CAD Generation and Reconstruction}
DeepCAD~\cite{wu2021deepcad} represents CAD models as sequences of sketch and extrusion commands and provides a large-scale dataset paired with construction histories. Based on this representation, subsequent methods reconstruct CAD sequences from point clouds~\cite{khan2024cadsignet}, text~\cite{khan2024textcad}, and images~\cite{chen2025cadcrafter,tan2026img2cadseq}. However, the statistical properties of the widely used DeepCAD parameterization and their effects on model learning have received limited attention. We identify a representation-induced bias in the DeepCAD parameterization: its local normalization assigns a substantial portion of geometric scale information to a separate scale factor $S$, resulting in highly concentrated distributions for several normalized geometric parameters. We therefore revise the representation to produce more balanced parameter distributions and reduce reliance on frequency priors.

\subsection{Sim2Real Generalization for Image-to-CAD Reconstruction}
Because paired real images and CAD command sequences are costly to collect, existing image-to-CAD methods are typically trained on synthetic renderings. The appearance gap between synthetic and real images, including differences in materials, textures, illumination, and imaging conditions, limits real-world generalization. CADCrafter~\cite{chen2025cadcrafter} promotes synthetic-to-real generalization by extracting texture-invariant geometric features from synthetic renderings, while Img2CADSeq~\cite{tan2026img2cadseq} introduces an intermediate point-cloud representation and contrastive alignment. In contrast, our method performs adaptation at both the image and feature levels by combining geometry-constrained synthetic-to-real translation with multi-positive contrastive learning across viewpoints and image domains.

\begin{figure*}[ht]
    \centering
\includegraphics[width=0.9\textwidth]{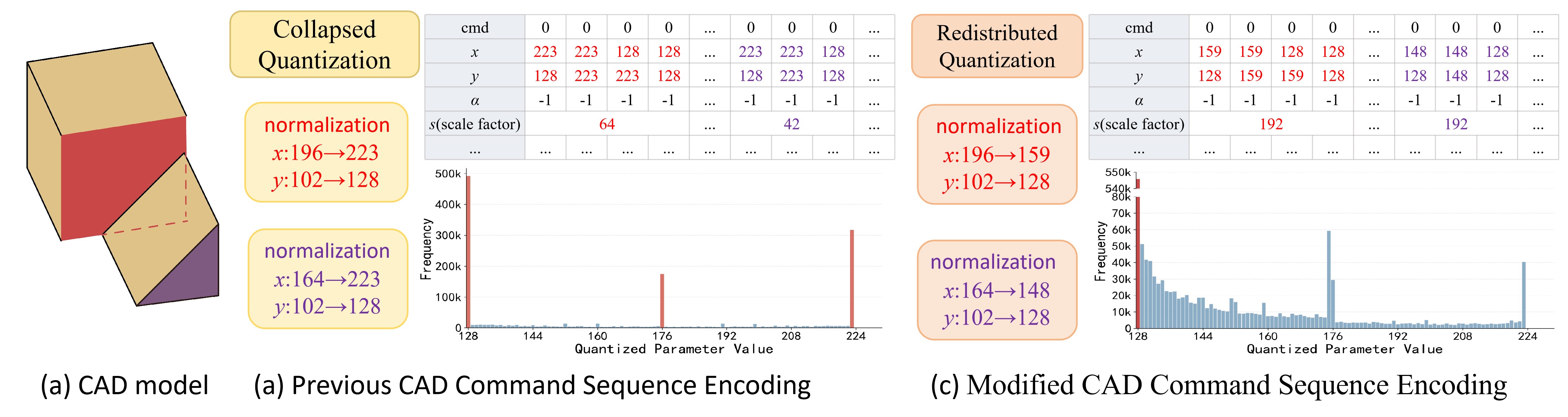}
    \caption{Revision of the DeepCAD parameter representation and its effect on parameter distributions: several parameters concentrate on very few valid values (\textit{e.g.}, most of $x$ and $y$ are concentrated at three specific values: 128, 176, and 223).} 
    \label{data_refinement}
        \vspace{-4mm}
\end{figure*}

\section{Benchmark Construction}
In this section, we first examine a previously underexplored distributional limitation in the parameter representation of DeepCAD and introduce the revised DeepCAD representation. We then present \textbf{OpenRealCAD}, a real-image benchmark consisting of separate subsets for model adaptation and evaluation. Together, the revised DeepCAD representation and OpenRealCAD provide a unified benchmark for training, adapting, and evaluating image-to-CAD reconstruction models under Sim2Real settings.

\subsection{Preliminaries: CAD Command Sequence Encoding}

A parametric CAD model is typically constructed through an ordered sequence of modeling operations. A common workflow begins by drawing a closed profile, referred to as a sketch, on a two-dimensional plane and subsequently extruding the profile along a specified direction to form a three-dimensional solid. More complex models can be constructed by sequentially creating multiple sketch-extrusion units and combining them through Boolean operations such as union, subtraction, and intersection. We refer to the commands and associated parameters that describe this construction process as a \emph{CAD command sequence}.

Although commercial CAD systems support a broad range of modeling operations, a compact set of commonly used commands, including sketching, extrusion, and Boolean operations, is sufficient to represent a wide variety of mechanical parts. To obtain a structured representation suitable for neural networks, we follow the sketch-extrusion formulation introduced by DeepCAD and encode each command and its associated parameters as a 17-dimensional vector:
$
\mathbf{v}_i =
[\mathrm{Command}, x, y, \alpha, f, r,
\theta, \phi, \gamma,
p_x, p_y, p_z,
s, e_1, e_2, b, u]
$
The first dimension specifies the command type, while the remaining dimensions store the corresponding geometric and operational parameters. Parameter entries that are not applicable to a given command are assigned a predefined invalid value. In addition to the modeling commands line, arc, circle, and extrusion, the representation contains two special tokens: $\langle \mathrm{SOL} \rangle$ (Start of Loop), which marks the beginning of a closed sketch profile, and $\langle \mathrm{EOS} \rangle$ (End of Sequence), which indicates the termination of the CAD command sequence.
The command type is defined as $C_i \in
\{
\langle \mathrm{SOL} \rangle,
L,
A,
C,
E,
\langle \mathrm{EOS} \rangle
\}
\label{eq}
$
and the parameters associated with each command, together with their meanings, are summarized in Table~\ref{tab:cad_command_parameters}, where $b$ specifies the Boolean relationship between the current extrusion and the existing solid. It takes values from $0$ to $3$, corresponding to the creation of a new solid, union, subtraction, and intersection, respectively. The parameter $u$ specifies the extrusion mode and takes values from $0$ to $2$, corresponding to one-sided, symmetric, and two-sided extrusion, respectively.

\subsection{Investigating Parameter Bias in DeepCAD}
We first examine the parameter distributions induced by the DeepCAD command-sequence representation. Although continuous parameters are quantized into 256 discrete values, several geometric parameters occupy only a small subset and are strongly concentrated around a few dominant values. Figure~\ref{data_refinement}(a) and (b) compare representative parameter distributions before and after our representation revision.

The concentration is most pronounced for sketch coordinates and radii. For $x$ and $y$, the values 128, 176, and 223 collectively account for 61.62\% of all observations, while radius values 47 and 48 account for 55.86\% of the samples. Rotation parameters are also concentrated at four values corresponding to the principal orthogonal orientations; however, this pattern largely reflects the discrete orientation structure of the underlying CAD models rather than the same normalization-induced bias. By comparison, translation and extrusion-distance parameters are more dispersed after excluding the zero value encoded by 128.

This concentration mainly arises from the local normalization adopted by DeepCAD. Each sketch-extrusion unit is independently rescaled to fit the quantization range, and the corresponding scaling ratio is stored in the scale parameter $s$. In multi-unit models, independently normalizing geometrically different units repeatedly maps their local extrema to similar discrete values, while transferring much of their relative size information from the geometric parameters to $s$.

Figure~\ref{data_refinement}(a) illustrates this effect using a composite model containing substructures with different dimensions. Under the original representation, geometrically distinct endpoints may be assigned identical quantized coordinates, while their relative dimensions are encoded primarily by $s$, resulting in concentrated geometric parameters and a comparatively dispersed scale distribution.

This information imbalance can affect both model learning and evaluation. A model may obtain high accuracy on strongly concentrated parameters by predicting dominant values rather than inferring geometry from the input image. At the same time, uniformly averaging parameter accuracies may understate errors in $s$, even though the scale factor carries substantial geometric information and controls the dimensions of an entire sketch--extrusion unit. Consequently, similar parameter accuracies do not necessarily imply comparable geometric fidelity.

To address this issue, we reparameterize all sketch-extrusion units in a shared scale space while preserving their relative dimensions. Specifically, we redistribute the scale information stored in $s$ to each affected geometric parameter. For each scale-dependent quantized parameter $p_{\mathrm{ori}}$, we absorb the unit-level scale factor $s$ into the parameter as
\begin{equation}
p_{\mathrm{new}}
=
128 + (p_{\mathrm{ori}}-128)\frac{s}{192},
\label{eq:scale_redistribution}
\end{equation}
where 128 denotes the quantized origin and 192 is the reference value corresponding to the maximum scale used in the original representation.

This transformation is equivalent to expressing every sketch-extrusion unit under a common reference scale and encoding its original relative size directly in the corresponding geometric parameters. The explicit scale parameter $s$ is therefore no longer required and is removed from the command vector. Because the transformation preserves the relative dimensions and spatial relationships of the original units, it retains the underlying CAD geometry while producing a more balanced parameterization.

We additionally canonicalize rotation and translation parameters. Since fewer than 0.01\% of the samples contain non-orthogonal rotations, we remove these rare cases and represent rotations using a unified set of orthogonal orientations. For translation, we set the first sketch-extrusion unit as the reference origin and express all subsequent unit translations relative to it. The fixed translation of the reference unit is excluded from parameter-accuracy evaluation.

We refer to the resulting representation as the revised DeepCAD representation. As shown in Figure~\ref{data_refinement}(b), after reparameterization, no individual value accounts for more than 3.5\% of either coordinate distribution, while the maximum frequency for the radius parameter decreases to 14.6\%. These results indicate that the revised DeepCAD representation substantially reduces parameter concentration while preserving the relative dimensions and spatial relationships of the retained CAD models.

\begin{table}[t]
\centering
\setlength{\tabcolsep}{2.5pt}

\begin{tabular}{@{} c r c >{\raggedright\arraybackslash}p{0.46\columnwidth} @{}}
\toprule
\textbf{Command}
& \multicolumn{3}{c}{\textbf{Parameters and Descriptions}} \\
\midrule

$\langle \mathrm{SOL} \rangle$
& \multicolumn{3}{c}{$\varnothing$} \\
\midrule

\begin{tabular}[c]{@{}c@{}}
$L$ \\
(Line)
\end{tabular}
& $x,y$
& :
& Endpoint coordinates \\
\midrule

\multirow{3}{*}{
\begin{tabular}[c]{@{}c@{}}
$A$ \\
(Arc)
\end{tabular}
}
& $x,y$    & : & Endpoint coordinates \\
& $\alpha$ & : & Sweep angle \\
& $f$      & : & Direction flag \\
\midrule

\multirow{2}{*}{
\begin{tabular}[c]{@{}c@{}}
$C$ \\
(Circle)
\end{tabular}
}
& $x,y$ & : & Center coordinates \\
& $r$   & : & Radius \\
\midrule

\multirow{6}{*}{
\begin{tabular}[c]{@{}c@{}}
$E$ \\
(Extrude)
\end{tabular}
}
& $\theta,\phi,\gamma$ & : & Plane rotation \\
& $p_x,p_y,p_z$        & : & Plane translation \\
& $s$                  & : & Scale factor \\
& $e_1,e_2$            & : & Extrusion distances \\
& $b$                  & : & Boolean type \\
& $u$                  & : & Extrusion mode \\
\midrule

$\langle \mathrm{EOS} \rangle$
& \multicolumn{3}{c}{$\varnothing$} \\
\bottomrule
\end{tabular}

\caption{CAD command types and their parameters.}
\label{tab:cad_command_parameters}
\end{table}

\subsection{Details of OpenRealCAD}
To support reproducible evaluation of Sim2Real image-to-CAD reconstruction, we construct \textbf{OpenRealCAD}, a real-image benchmark with paired ground-truth CAD command sequences. We randomly select 392 source models from DeepCAD and re-encode their construction sequences using the revised DeepCAD representation. This ensures that the synthetic and real domains share the same output parameterization and avoids representation mismatch during cross-domain training and evaluation.

We fabricate each model through 3D printing and capture it from four predefined viewpoints under controlled backgrounds and illumination. This setup limits pose and scene variation, enabling direct comparison between synthetic renderings and real photographs of the same geometry.

OpenRealCAD is intended as a controlled real-image benchmark rather than an unconstrained in-the-wild dataset. By limiting background clutter, severe occlusion, and large viewpoint changes, the benchmark focuses evaluation on appearance differences such as materials, surface textures, illumination, shadows, and imaging artifacts. It therefore provides an intermediate setting between synthetic-only evaluation and deployment in unconstrained real environments.

We split OpenRealCAD into disjoint training and test subsets at the CAD-instance level using a 7:3 ratio. To prevent instance-level leakage, all synthetic renderings associated with real-test models are excluded from synthetic training. The real training subset supports limited real-domain adaptation, while the held-out test subset evaluates generalization to unseen physical objects. Our dataset will be publicly released upon acceptance.

\section{Method}
\begin{figure*}[ht]
    \centering
\includegraphics[width=0.9\textwidth]{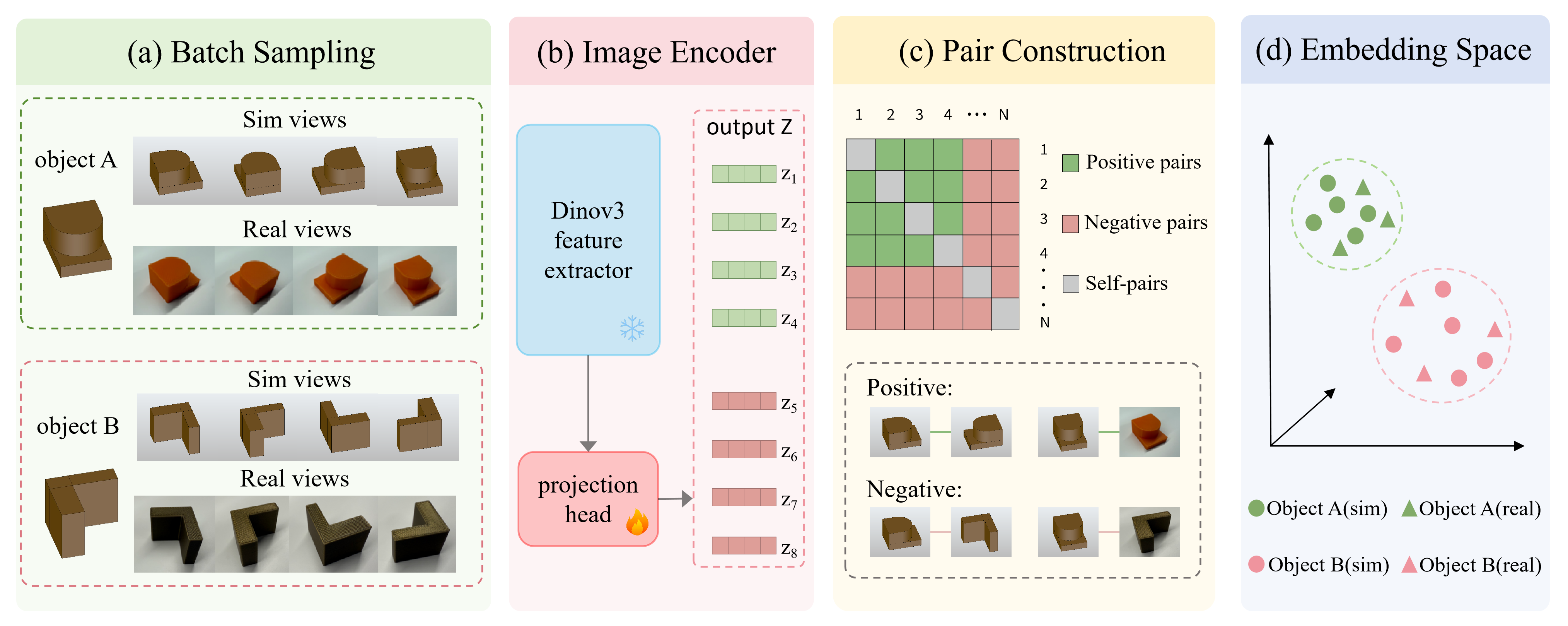}
    \caption{\textbf{Overview of the feature-learning pipeline in RealCAD.} Multi-view images from the synthetic and real domains are encoded by a frozen pretrained DINOv3 encoder and projected into a shared representation space, where multi-positive contrastive learning promotes cross-view and cross-domain consistency for the same CAD model.}
    \label{model}
        \vspace{-4mm}
\end{figure*}

Building upon the revised DeepCAD representation, we propose \textbf{RealCAD}, an image-to-CAD reconstruction framework for Sim2Real generalization. RealCAD mitigates the synthetic-to-real discrepancy at both the image and feature levels. At the image level, geometry-constrained synthetic-to-real translation reduces appearance differences while preserving the underlying object structure. At the feature level, multi-positive contrastive learning promotes consistent representations of the same CAD model across viewpoints and image domains. Figure~\ref{model} illustrates the feature-learning pipeline, including multi-view sampling, visual encoding, and cross-view and cross-domain representation alignment. The learned representations are subsequently decoded by a Transformer decoder into CAD command sequences and their associated parameters.

\subsection{Geometry-Constrained Synthetic-to-Real Translation}

To exploit abundant synthetic data while reducing the appearance gap to real photographs, we translate synthetic renderings toward the real-image domain. However, unconstrained image translation may alter object geometry and introduce inconsistent supervision for CAD reconstruction. We therefore employ an edge-conditioned ControlNet~\cite{zhang2023adding} to constrain the translation process. 
Let \(x_{m,v}^{\mathrm{syn}}\) denote the synthetic rendering of CAD model
\(m\) from viewpoint \(v\), and let \(e_{m,v}\) be its corresponding edge map.
The translated image is generated as
\begin{equation}
x_{m,v}^{\mathrm{tr}}
=
G_{\mathrm{ctrl}}
\left(
x_{m,v}^{\mathrm{syn}},
e_{m,v},
p;
\boldsymbol{\epsilon}
\right),
\label{eq:controlnet_translation}
\end{equation}
where \(G_{\mathrm{ctrl}}\) denotes the edge-conditioned ControlNet, \(p\)
is the text prompt, and \(\boldsymbol{\epsilon}\) denotes the random noise
used during diffusion sampling.
The edge condition encourages the translated images to preserve object contours while adapting domain-specific appearance attributes such as materials, textures, illumination, and shadows. By retaining the underlying geometry while making synthetic appearances more similar to real photographs, the translated images narrow the image-level domain gap and facilitate Sim2Real generalization.

\subsection{Multi-Positive Contrastive Learning}
\label{sec:multi_positive_contrastive}

For each CAD model, we render four fixed-view images and extract visual features using a pretrained DINOv3-ViT-7B encoder~\cite{simeoni2025dinov3}, whose parameters are kept frozen throughout training. A learnable projection layer maps the features into a shared latent space. The resulting representation is used both for contrastive alignment and as input to the Transformer decoder. Since different views and image-domain variants of the same CAD model share identical underlying geometry, we treat them as positive samples and regard images from different models as negatives.

For samples $(i,j)$, let $\hat{\mathbf{z}}_i$ and $\hat{\mathbf{z}}_j$ denote the $\ell_2$-normalized latent representations. The similarity logit is defined as
\begin{equation}
s_{ij}
=
\alpha
\hat{\mathbf{z}}_i^{\top}
\hat{\mathbf{z}}_j
+
b,
\qquad
\alpha=\exp(t),
\label{eq:pair_similarity}
\end{equation}
where \(t\) and \(b\) are learnable log-scale and bias parameters, respectively. This parameterization guarantees a positive logit scale \(\alpha>0\).

We set $y_{ij}=1$ if the two samples are associated with the same CAD model and $y_{ij}=-1$ otherwise. The multi-positive contrastive loss is formulated as
\begin{equation}
\mathcal{L}_{\mathrm{siglip}}
=
-\frac{1}{|\Omega|}
\sum_{(i,j)\in\Omega}
\log \sigma\left(y_{ij}s_{ij}\right),
\label{eq:siglip_loss}
\end{equation}
where $\Omega$ denotes the set of valid non-self sample pairs and $\sigma(\cdot)$ is the sigmoid function. Because each CAD model may appear under multiple viewpoints and image domains, each sample can have multiple positives. This objective promotes cross-view and cross-domain consistency in the feature space while separating representations of different CAD models.

\subsection{CAD Sequence Reconstruction}

Each latent representation is independently passed to a Transformer decoder to reconstruct the complete CAD command sequence. At each sequence position, the decoder predicts the command type and its associated discretized parameters. The reconstruction objective is defined as
\begin{equation}
\mathcal{L}_{\mathrm{rec}}
=
\lambda_{1}\mathcal{L}_{\mathrm{cmd}}
+
\lambda_{2}\mathcal{L}_{\mathrm{para}},
\label{eq:rec_loss_components}
\end{equation}
where \(\mathcal{L}_{\mathrm{cmd}}\) and \(\mathcal{L}_{\mathrm{para}}\) denote the cross-entropy losses for command-type classification and discretized parameter classification, respectively, and \(\lambda_{1}\) and \(\lambda_{2}\) balance their contributions.
The overall training objective is
\begin{equation}
\mathcal{L}
=
\mathcal{L}_{\mathrm{siglip}}
+
\mathcal{L}_{\mathrm{rec}},
\label{eq:total_loss}
\end{equation}
which jointly optimizes cross-view and cross-domain representation alignment and CAD sequence reconstruction.

\section{Experiments}

\begin{figure}[ht]
    \centering
\includegraphics[width=0.45\textwidth]{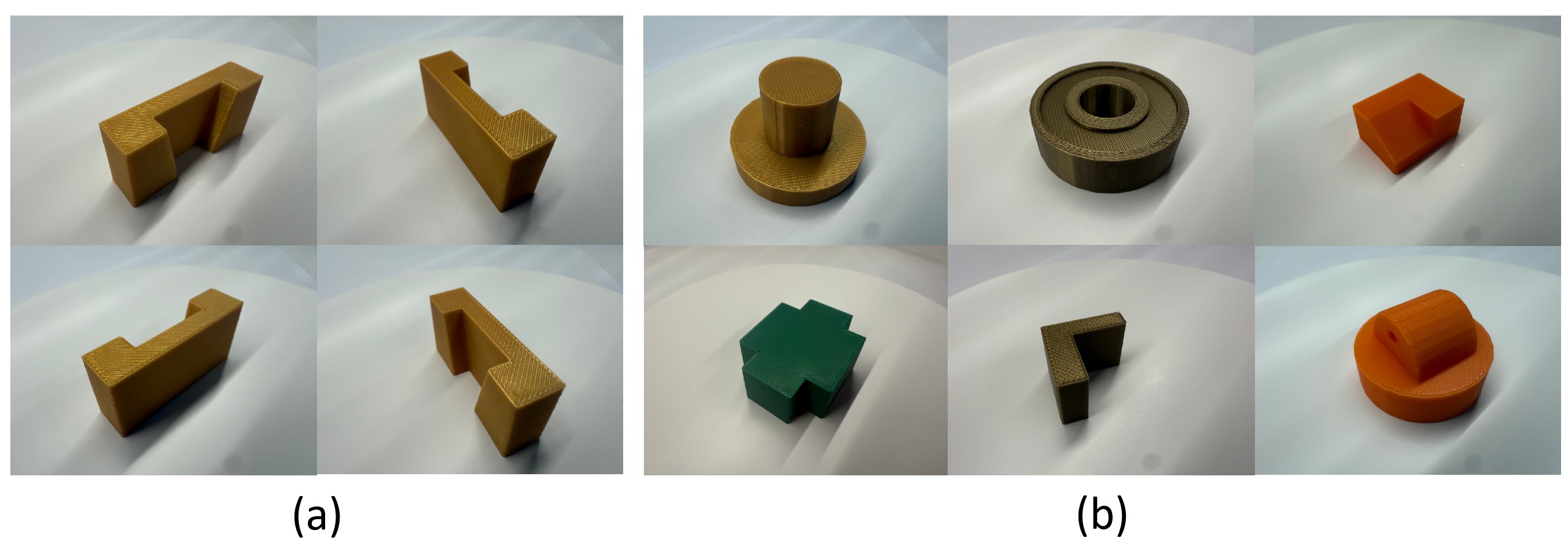}
    \caption{Examples from our OpenRealCAD dataset: (a) multi-view images of 3D-printed CAD models, and (b) additional 3D-printed models captured using smartphones.} 
    \label{fig:realcaddata}
        \vspace{-4mm}
\end{figure}

\subsection{Datasets Setup}
We train and evaluate all models using three types of data: CAD command sequences encoded with the revised DeepCAD representation, synthetic images rendered from these command sequences, and real images collected in OpenRealCAD. Figure~\ref{fig:realcaddata} presents representative examples from OpenRealCAD, showcasing multiple physical CAD models captured from different viewpoints. For each CAD model, we render synthetic images from four fixed viewpoints using OpenCASCADE. The synthetic data are split into training and test sets at a ratio of 9:1, while OpenRealCAD is divided at the CAD-instance level into real training and test sets at a ratio of 7:3. To prevent instance leakage, synthetic renderings of models in the real test set are excluded from training. We additionally generate edge-conditioned translated images for image-level adaptation.

\subsection{Implementation Details}

We use a pretrained DINOv3-ViT-7B encoder to extract a 4096-dimensional feature from each image. All models are trained for 300 epochs with a batch size of 128 and an initial learning rate of $1\times10^{-3}$ on eight NVIDIA A100 GPUs. Unless otherwise specified, all methods use the same data splits and evaluation protocol.

\subsection{Baselines}

We compare RealCAD with DeepCAD, Img2CAD, and CADCrafter. Since the original DeepCAD framework does not support image-conditioned reconstruction, we augment it with the same pretrained DINOv3 encoder used in RealCAD, enabling CAD sequence reconstruction from image features. For Img2CAD, we implement the method following its published architecture and training objective. For CADCrafter, we reproduce its supervised image-to-CAD reconstruction pipeline but omit the DPO stage. All methods are trained and evaluated using the same data splits and metrics, while unspecified implementation settings are kept consistent whenever possible.

\subsection{Experimental Results}
Following prior image-to-CAD studies, we evaluate performance using Command Accuracy (Cmd.), Parameter Accuracy (Para.), median Chamfer Distance (mCD), and Invalidity Ratio (IR). Cmd. measures command-type accuracy at individual sequence positions, while Para. measures classification accuracy over valid command parameters. For successfully constructed predictions, mCD quantifies the geometric discrepancy between the reconstructed and ground-truth models. IR denotes the proportion of predicted sequences that cannot be parsed or constructed by the CAD kernel. Higher Cmd. and Para. values are better, whereas lower mCD and IR values are preferred.

\noindent\textbf{Effect of DeepCAD Representation Refinement.}
To quantify the extent to which parameter accuracy can be obtained from dataset-frequency priors without using visual
evidence, we compare the trained reconstruction model with a frequency-only baseline. For each parameter type, the baseline always predicts the most frequent nonzero quantized value observed in the training set. The zero-valued bin 128 is excluded from both prior estimation and evaluation because it represents a legitimate geometric zero and remaines unaffected by our reparameterization. The accuracy of the frequency-only baseline therefore provides an estimate of the performance attainable solely from the marginal parameter distribution. We train the same reconstruction architecture separately using the original DeepCAD representation and the revised DeepCAD representation.

\begin{table}[t]
\centering
\setlength{\tabcolsep}{3.2pt}
\renewcommand{\arraystretch}{1.05}
\small

\begin{tabular}{lccc@{\hspace{5pt}}ccc}
\toprule
\multirow{2}{*}{\textbf{Parameter}}
& \multicolumn{3}{c}{\textbf{DeepCAD}}
& \multicolumn{3}{c}{\textbf{Revised DeepCAD}} \\
\cmidrule(lr){2-4}
\cmidrule(lr){5-7}
& \textbf{Model}
& \textbf{Prior}
& $\boldsymbol{\Delta}$
& \textbf{Model}
& \textbf{Prior}
& $\boldsymbol{\Delta}$ \\
\midrule
Line
& 57.71 & 16.77 & 40.94
& 50.27 &  3.14 & 47.13 \\

Arc
& 51.44 & 31.31 & 20.13
& 50.61 & 22.93 & 27.68 \\

Circle
& 72.23 & 26.13 & 46.10
& 57.81 & 6.41 & 51.40 \\

Scale ($s$)
& 32.64 & 11.31 & 21.33
& -- & -- & -- \\

Rotation
& 88.48 & 54.15 & 34.33
& 87.05 & 54.15 & 32.90 \\

Translation
& 50.07 & 7.15 & 42.92
& 70.34 & 1.64 & 68.70 \\

Extrusion
& 34.38 & 14.78 & 19.60
& 33.07 & 14.78 & 18.29 \\
\bottomrule
\end{tabular}
\caption{
Comparison of the original DeepCAD representation and the revised DeepCAD representation. \textit{Model} denotes the accuracy of the trained reconstruction model, while \textit{Prior} denotes the accuracy obtained by always predicting the most frequent valid training-set value for each parameter type. $\Delta=\textit{Model}-\textit{Prior}$ measures the gain over the frequency-only baseline. All values are reported as percentages.
}
\label{tab:dataset_refinement}
\end{table}

As shown in Table~\ref{tab:dataset_refinement}, the frequency-only baseline achieves nontrivial accuracy under the original DeepCAD representation, particularly for Line, Arc, and Circle parameters, indicating that a considerable portion of parameter accuracy can be obtained without visual evidence. The revised  DeepCAD representation substantially reduces the prior accuracy for Line, Circle, and Translation from 16.77\%, 26.13\%, and 7.15\% to 3.14\%, 6.41\%, and 1.64\%, respectively, while retaining competitive model accuracy and enlarging the corresponding gains over the prior baseline. Rotation and Extrusion remain nearly unchanged because their distributions are not directly affected by the proposed scale redistribution. These results show that the revised DeepCAD representation reduces the accuracy attainable from marginal frequency priors and makes parameter accuracy more indicative of image-conditioned geometric inference.

\begin{table}[t]
\centering

\begin{tabularx}{\columnwidth
}{
@{}l
>{\centering\arraybackslash}X
>{\centering\arraybackslash}X
>{\centering\arraybackslash}X
>{\centering\arraybackslash}X@{}
}
\toprule
\textbf{Method}
& \textbf{Cmd. $\uparrow$}
& \textbf{Para. $\uparrow$}
& \textbf{mCD $\downarrow$}
& \textbf{IR $\downarrow$} \\
\midrule

\multicolumn{5}{@{}l}{\textbf{Synthetic-Domain Test}} \\
\midrule
DeepCAD$^*$
& 69.63 & 54.52 & \textbf{0.207} & \textbf{15.02} \\

Img2CAD$^{\dagger}$
& 73.02 & 57.20 & 0.226 & 23.02 \\

CADCrafter$^{\dagger}$
& 67.82 & 45.80 & 0.246 & 22.92 \\

\textbf{Ours}
& \textbf{72.94}
& \textbf{61.35}
& 0.220
& 26.85 \\

\midrule
\multicolumn{5}{@{}l}{\textbf{Real-Domain Test}} \\
\midrule
DeepCAD$*$
& 43.34 & 38.35 & \textbf{0.293} & \textbf{15.38} \\

Img2CAD$^{\dagger}$
& 61.80 & 43.08 & 0.302 & 30.77 \\

CADCrafter$^{\dagger}$
& 57.36 & 37.30 & 0.309 & 17.09 \\

\textbf{Ours}
& \textbf{64.85}
& \textbf{52.03}
& 0.308
& 48.72 \\
\bottomrule
\end{tabularx}

\caption{{Quantitative comparison on the synthetic- and real-domain test sets. $^{*}$ indicates our image-conditioned adaptation of the official DeepCAD code with a DINOv3 encoder. $^{\dagger}$ indicates our reproduction based on the published paper, as the official code is unavailable. CADCrafter is reproduced without the DPO stage.}}
\label{tab:main_comparison}
\end{table}

\noindent\textbf{Main Results.}Table~\ref{tab:main_comparison} shows that RealCAD achieves the best token-level performance on the real-domain test set, improving command and parameter accuracies over CADCrafter by 7.49 and 14.73 percentage points, respectively, supporting the effectiveness of combining image-level geometry-constrained synthetic-to-real translation with feature-level multi-positive contrastive learning for reducing the Sim2Real domain gap.  However, these gains do not translate into a comparable improvement in mCD and are accompanied by a higher invalid ratio, since token-level metrics weight individual predictions similarly, whereas the final geometric quality is often dominated by a small number of geometry-sensitive parameters or structural operations. CAD parameters have highly heterogeneous geometric sensitivities: correcting a large number of low-impact parameters can improve token-level accuracy without noticeably changing the reconstructed shape, whereas a single error in a topology-defining command or a geometry-sensitive parameter can dominate the final reconstruction error. Moreover, CAD execution is sequential and compositional, such that early prediction errors may alter the reference frame or feasible domain of subsequent operations. Therefore, geometry quality is determined not only by the number of correctly predicted parameters, but also by their semantic importance, cross-command consistency, and executability as a complete program.

\noindent\textbf{Ablation Studies.}
We evaluate geometry-constrained synthetic-to-real translation (ST) and multi-positive contrastive learning (CL) by independently removing each component. All variants use identical data splits, architectures, and optimization settings. 

\begin{table}[t]
\centering
\setlength{\tabcolsep}{4pt}

\begin{tabular}{cccccc}
\toprule
\multicolumn{2}{c}{\textbf{Modules}}
& \multicolumn{2}{c}{\textbf{Synthetic Images (\%)}}
& \multicolumn{2}{c}{\textbf{Real Images (\%)}} \\
\cmidrule(lr){1-2}
\cmidrule(lr){3-4}
\cmidrule(lr){5-6}
\textbf{CL}
& \textbf{ST}
& \textbf{Cmd.}
& \textbf{Para.}
& \textbf{Cmd.}
& \textbf{Para.} \\
\midrule

$\times$
& $\times$
& \textbf{75.98} & \textbf{62.21}
& 60.18          & 49.98 \\

$\checkmark$
& $\times$
& 75.98 & 62.15
& 61.65 & 48.98 \\

$\times$
& $\checkmark$
& 72.51 & 60.90
& 62.61 & 51.48 \\

$\checkmark$
& $\checkmark$
& 72.94          & 61.35
& \textbf{64.85} & \textbf{52.03} \\

\bottomrule
\end{tabular}

\caption{Effects of multi-positive contrastive learning (CL) and
geometry-constrained synthetic-to-real translation (ST) on CAD reconstruction.
Cmd. and Para. denote command and parameter accuracy, respectively.
Bold values indicate the best result in each column.}
\label{tab:ablation}
\end{table}

As shown in Table~\ref{tab:ablation}, CL alone increases real-domain command accuracy from 60.18\% to 61.65\%, although parameter accuracy decreases slightly. ST alone improves command and parameter accuracy to 62.61\% and 51.48\%, indicating that ST provides a larger individual gain than CL under this setting. Combining both modules yields the best real-domain results of 64.85\% and 52.03\%, improving over the baseline by 4.67 and 2.05 percentage points, respectively. The combined results suggest that ST and CL provide complementary benefits. ST narrows the appearance gap, while CL further promotes cross-view and cross-domain feature consistency.

The complete model shows a modest decrease in synthetic-domain command accuracy while maintaining comparable parameter accuracy. This result may reflect a trade-off between fitting synthetic-domain appearances and improving generalization to real images, which is consistent with the Sim2Real objective of RealCAD.

\noindent\textbf{Qualitative Results.} Detailed comparisons of the qualitative results will be provided in the supplementary material.


\subsection{Limitations}
Although this paper identifies and systematically analyzes several issues in recent CAD-related research, the current framework still has two main limitations.
First, although the revised DeepCAD representation reduces parameter concentration caused by local normalization, CAD parameters remain intrinsically long-tailed because coordinates and geometric entities near the sketch origin occur more frequently. Long-tail-aware training and evaluation may further improve performance. Second, RealCAD mainly optimizes token-level accuracy without explicitly modeling CAD grammar, command dependencies, or geometric feasibility. Consequently, token-level gains do not always yield comparable improvements in geometric quality or program validity, as indicated by the modest mCD improvement and higher IR. Future work could incorporate grammar-constrained decoding and geometry- or execution-aware objectives.

\section{Conclusion}
In this paper, we focus on the robustness and reliability of Image-to-CAD methods in real-world application scenarios. We identify a previously overlooked parameter bias in the DeepCAD representation and introduce a revised parameterization that reduces geometric parameter concentration and frequency-based shortcuts. We also propose RealCAD, a Sim2Real image-to-CAD reconstruction framework combining geometry-constrained image translation with multi-positive contrastive learning to improve cross-domain and cross-view feature consistency. Experiments show that the revised representation reduces dependence on marginal frequency priors, while RealCAD improves command and parameter accuracy on real images. However, the limited gains in geometric quality and program validity suggest that token-level optimization alone is insufficient, motivating future work on geometry- and validity-aware learning.

\section{Acknowledgement}
\label{sec:acknowledgement}

This work was supported by the Fundamental Research Funds for the Central Universities at Tongji University under Grant No.~22120260376.

\nocite{*}

\bibliography{aaai2027}


\newpage

\appendix

\section{Supplementary Material}
This supplementary material provides additional analysis and qualitative evaluations 
to complement the main paper. We first investigate the parameter distribution bias 
in the DeepCAD dataset and provide detailed statistics for different CAD parameters. 
We further present qualitative reconstruction results under both synthetic and 
real-world image inputs.

\section{Detailed Analysis of the DeepCAD Dataset}


The DeepCAD dataset contains noticeable distribution biases among several 
geometric parameters in its original CAD representation. Such biases may lead 
models to overfit frequent parameter values and reduce their ability to infer 
accurate geometric attributes from input images. To provide a detailed analysis, 
we report the observed frequency and percentage of dominant parameter values 
in the following tables.

\begin{table}[h]
\centering

\begin{tabular}{ccc}
\toprule
Value & $x$ & $y$ \\
\midrule
128    & 492273 (32.27\%) & 667450 (43.77\%) \\
176    & 174920 (11.47\%) & 27623 (1.82\%) \\
223    & 317427 (20.82\%) & 199585 (13.09\%) \\
Others & 317427 (35.44\%) & 199585 (41.32\%) \\
\bottomrule
\end{tabular}
\caption{Distribution bias of parameters $x$ and $y$ in the DeepCAD dataset.}
\label{tab:xy_bias}
\end{table}

\begin{table}[h]
\centering
\begin{tabular}{cc}
\toprule
Value & $r$ \\
\midrule
47     & 22448 (11.62\%) \\
48     & 85071 (44.25\%) \\
Others & 84817 (44.12\%) \\
\bottomrule
\end{tabular}
\caption{Distribution bias of parameter $r$ in the DeepCAD dataset.}
\label{tab:r_bias}
\end{table}

\begin{table}[t]
\centering

\resizebox{\linewidth}{!}{
\begin{tabular}{c|ccc}
\toprule
Value 
& $\theta$ 
& $\phi$ 
& $\gamma$ \\
\midrule
64 
& 0 / 0\% 
& 145173 / 39.60\% 
& 14 / 0.00\% \\

128 
& 165852 / 45.21\%
& 213205 / 58.18\%
& 165913 / 45.28\% \\

192
&191242 / 52.15\%
&7842 / 2.15\%
&191791 /52.29\%\\

255
&8817 /2.43\%
&0 /0\%
&8847 /2.43\%

\\

Others
&737 /0.21\%
&428 /0.12\%
&83 /0.05\%

\\
\bottomrule
\end{tabular}
}
\caption{Distribution bias of rotation parameters 
$\theta$, $\phi$, and $\gamma$ in the DeepCAD dataset.}
\label{tab:rotation_bias}
\end{table}

\begin{table}[t]
\centering

\resizebox{\linewidth}{!}{
\begin{tabular}{c|ccc}
\toprule
Value
& $p_x$
& $p_y$
& $p_z$
\\
\midrule

32
&54523 /14.84\%
&15192 /4.16\%
&7859 /2.15\%

\\

128
&65157 /17.75\%
&190089 /51.87\%
&206149 /56.24\%

\\

Others
&247450 /67.41\%
&161138 /43.97\%
&152522 /41.61\%

\\

\bottomrule
\end{tabular}
}
\caption{Distribution bias of translation parameters 
$p_x$, $p_y$, and $p_z$ in the DeepCAD dataset.}
\label{tab:translation_bias}
\end{table}

The above statistics demonstrate that the original DeepCAD representation
suffers from substantial distribution biases across different types of CAD
parameters, including coordinates, geometric sizes, rotations, and translations.

For the coordinate parameters $x$ and $y$, the distributions are highly
imbalanced toward several specific values. In particular, the value 128
accounts for 32.27\% of $x$ occurrences and 43.77\% of $y$ occurrences,
respectively. Meanwhile, several other discrete values, such as 176 and 223,
also appear frequently. This indicates that the sketch generation process in
DeepCAD tends to favor specific coordinate locations rather than uniformly
covering the spatial domain. Such bias may encourage models to rely on
frequent coordinate priors instead of accurately inferring spatial positions
from visual inputs.

The radius parameter $r$ also exhibits a noticeable imbalance. The value 48
alone appears in 44.25\% of all samples, while value 47 contributes another
11.62\%. Together, these two values account for more than half of the observed
radius distributions. This concentration suggests that the dataset contains a
strong preference toward specific object scales, potentially limiting the
model's ability to generalize to objects with unseen or uncommon geometric
sizes.

\begin{figure*}[h]
    \centering
\includegraphics[width=0.9\textwidth]{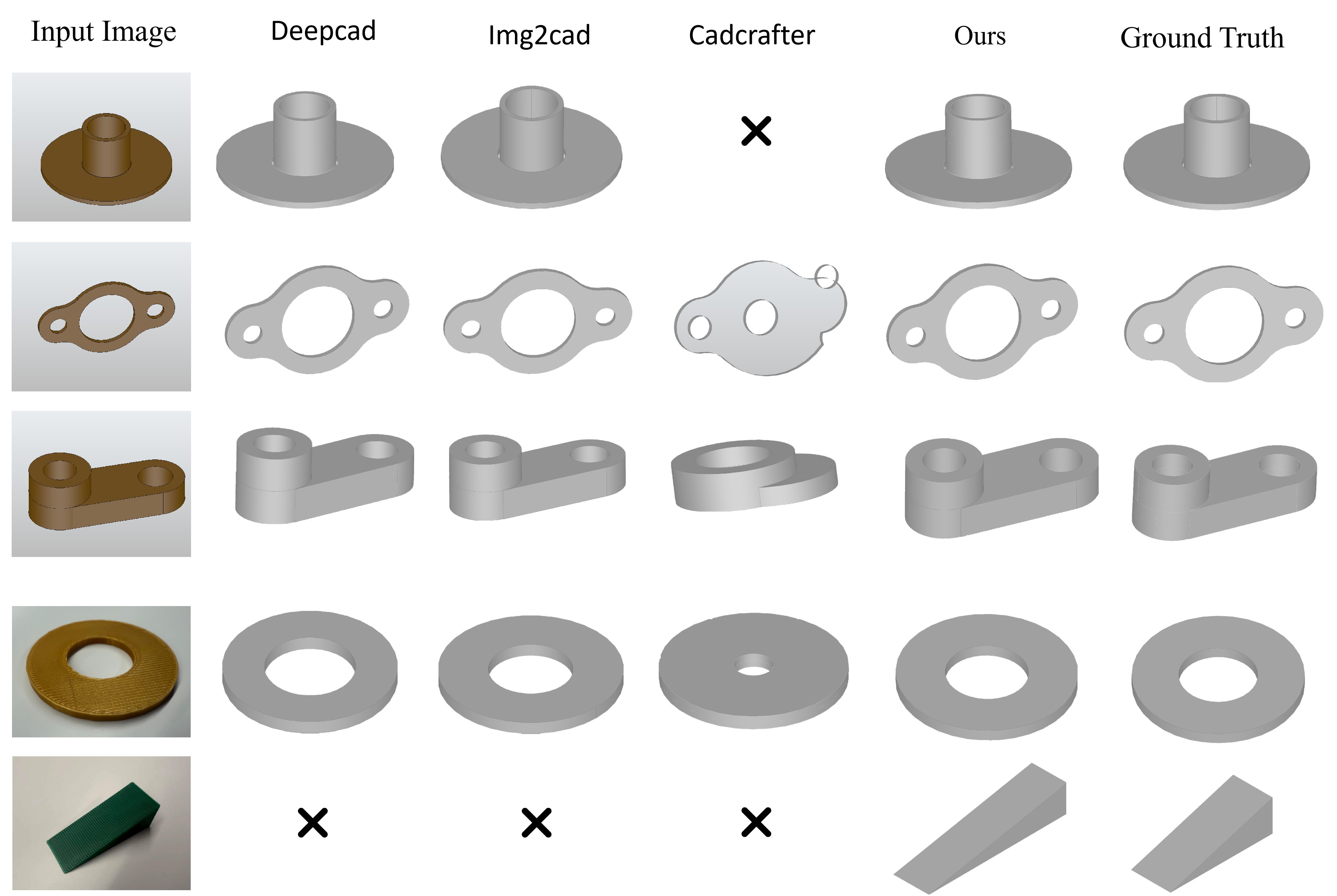}
\caption{Qualitative comparison of image-to-CAD reconstruction results on
synthetic and real-world image inputs. Our method produces more accurate
geometric structures and dimensions compared with existing approaches,
especially for complex objects under real-world conditions.}
    \label{fig:qualitative}
        \vspace{-4mm}
\end{figure*}

The rotation parameters $\theta$, $\phi$, and $\gamma$ demonstrate even
stronger concentration patterns. For $\theta$, the values 128 and 192 account
for 45.21\% and 52.15\% of occurrences, respectively, covering 97.36\% of all
samples. Similarly, $\gamma$ is dominated by values 128 and 192, which
together represent 97.57\% of the distribution. For $\phi$, values 64 and 128
occupy 39.60\% and 58.18\% of samples, respectively. These results indicate
that the original dataset contains only limited diversity in object
orientations, causing the rotation representation to be strongly biased toward
a few canonical poses.

The translation parameters $p_x$, $p_y$, and $p_z$ also exhibit considerable
spatial bias. The value 128 dominates both $p_y$ and $p_z$, accounting for
51.87\% and 56.24\% of samples, respectively. In contrast, $p_x$ shows a
relatively broader distribution, where the ``Others'' category contributes
67.41\% of occurrences. This suggests that although horizontal translations
are relatively diverse, vertical and depth translations remain concentrated
around specific locations.

Overall, these observations reveal that the original DeepCAD representation
contains strong prior biases across multiple parameter dimensions. Such
imbalanced distributions may introduce unintended shortcuts during training,
where models can exploit frequent parameter patterns instead of learning
generalizable geometric reasoning from image inputs. Therefore, addressing
parameter distribution bias is important for improving robustness and
generalization in image-to-CAD reconstruction.

\section{Qualitative Results}

Figure~\ref{fig:qualitative} presents qualitative comparisons between our method 
and several reproduced baselines under both synthetic and real-world image inputs.

For synthetic images, our method successfully reconstructs all provided examples 
with accurate geometric structures. Existing approaches, including DeepCAD and 
Img2CAD, achieve comparable reconstruction quality on synthetic inputs. However, 
CadCrafter occasionally produces inaccurate dimensions and spatial locations, 
resulting in incomplete reconstruction results.

For real-world image inputs, DeepCAD, Img2CAD, and CadCrafter are able to recover 
the overall topology of simple objects such as ring-like structures. Nevertheless, 
their predicted geometric dimensions are less accurate compared with our method. 
In particular, our approach is the only method capable of accurately reconstructing 
the triangular prism example, demonstrating stronger robustness when handling 
images with complex geometric structures and realistic variations.

\end{document}